\documentclass[runningheads]{llncs}

\usepackage{eccv}

\usepackage{eccvabbrv}

\usepackage{graphicx}
\usepackage{booktabs}
\usepackage{paralist}
\usepackage{multirow} 
\usepackage{tabularx}
\usepackage{tabulary}
\usepackage{comment}
\usepackage{amsmath}
\usepackage{amssymb}
\usepackage{animate}
\usepackage{xspace}
\usepackage{kotex}
\usepackage{pifont}
\usepackage{ulem}
\usepackage{subcaption}
\usepackage{rotating} % For vertical text rotation
\usepackage{xcolor}
\definecolor{grayblue}{RGB}{35,128,219}
\definecolor{navy1}{RGB}{17,34,163}

\usepackage[accsupp]{axessibility}  % Improves PDF readability for those with disabilities.

\usepackage{hyperref}

\usepackage{orcidlink}

\newcommand\norm[1]{\lVert#1\rVert}
\def\ours{DeblurGS\xspace}
\def\projmat{\mathbf{P}}
\def\rendermat{\mathcal{R}}
\def\bezier{B\'{e}zier\xspace}

\begin{document}

% ---------------------------------------------------------------
\title{DeblurGS: Gaussian Splatting for Camera Motion Blur} 

\author{Jeongtaek Oh\inst{1} \and
Jaeyoung Chung\inst{2} \and
Dongwoo Lee\inst{2} \and Kyoung Mu Lee\inst{1,2}}

\authorrunning{J. Oh et al.}

\institute{IPAI, Seoul National University, Korea \and Dept. of ECE \& ASRI, Seoul National University, Korea \\
\email{\{ohjtgood, robot0321, dongwoo.lee, kyoungmu\}@snu.ac.kr}
}
\maketitle

\begin{center}
    \centering
    \captionsetup{type=figure}
    \includegraphics[width=0.88\textwidth]{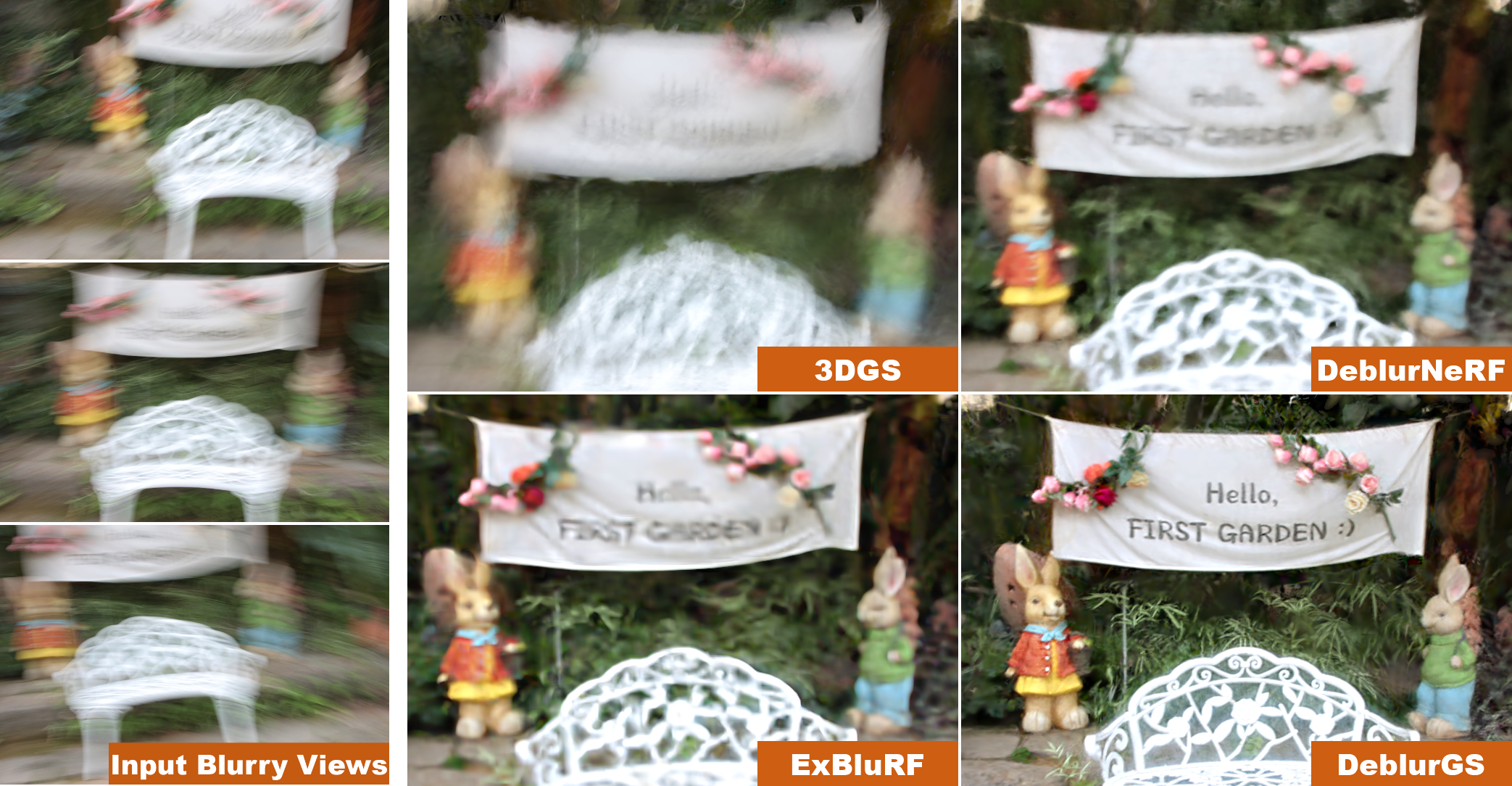}
    \captionof{figure}{\textbf{Novel View Synthesis with Blurry Views.} 
    Our \ours achieves state-of-the-art performance in novel view synthesis and deblurring compared to previous approaches.}
    
    \label{fig:teaser}
\end{center}%

\begin{abstract}
% %
\vspace{-0.3cm}

Although significant progress has been made in reconstructing sharp 3D scenes from motion-blurred images, a transition to real-world applications remains challenging.
The primary obstacle stems from the severe blur which leads to inaccuracies in the acquisition of initial camera poses through Structure-from-Motion, a critical aspect often overlooked by previous approaches.
To address this challenge, we propose DeblurGS, a method to optimize sharp 3D Gaussian Splatting from motion-blurred images, even with the noisy camera pose initialization.
We restore a fine-grained sharp scene by leveraging the remarkable reconstruction capability of 3D Gaussian Splatting. 
Our approach estimates the 6-Degree-of-Freedom camera motion for each blurry observation and synthesizes corresponding blurry renderings for the optimization process.
Furthermore, we propose Gaussian Densification Annealing strategy to prevent the generation of inaccurate Gaussians at erroneous locations during the early training stages when camera motion is still imprecise.
Comprehensive experiments demonstrate that our DeblurGS achieves state-of-the-art performance in deblurring and novel view synthesis for real-world and synthetic benchmark datasets, as well as field-captured blurry smartphone videos. 

\vspace{-0.2cm}

\keywords{3D Gaussian Splatting \and Camera Motion Deblurring  }
\end{abstract}

\section{Introduction}
\label{sec:intro}

3D reconstruction and photorealistic novel view synthesis are longstanding interests for industrial applications, such as AR/VR, autonomous driving, large scene reconstruction, and robot navigation.
Neural radiance fields (NeRF)~\cite{mildenhall2020nerf} and its numerous following works~\cite{barron2022mip,deng2022fov,xiangli2022bungeenerf,adamkiewicz2022vision} achieve great progress in such 3D vision tasks.
The practical deployment of NeRFs often involves cumbersome processes of capturing a number of individual images from different viewpoints.
Therefore, recording scenes through video is a more applicable and user-friendly setting.
However, multi-view input data acquired from video capturing is usually accompanied by camera motion blur, and the 3D reconstruction quality can be compromised by the blurry inputs.

We investigate the problem of novel view synthesis from motion blurred images.
Several works~\cite{lee2023exblurf, wang2023badnerf, Ma_deblurnerf, Lee_2023_CVPR} attempt to reconstruct 3D scenes from blurry multi-view images by introducing deblur modules to NeRF's framework. 
Although the aforementioned works demonstrate significant improvements in sharp reconstruction of scenes, they rely on the assumption that the exact poses of each image are known through the use of Blender~\cite{blender} or a beam-splitter camera~\cite{rim_2020_realblurdataset}.
However, this assumption is unrealistic, as Structure-from-Motion (SfM) with blurry images estimates camera poses with large errors.

More recently, 3D Gaussian Splatting (3DGS)~\cite{kerbl20233d} has attracted attention because of its real-time and high-quality rendering with fast training time.
While original 3DGS is designed to reconstruct 3D scenes from a set of clean images, we also find it an attractive option for restoring sharp scenes from blurry images because of its exceptional capability for fine-detailed reconstruction.
Despite the potential of 3DGS to capture fine-grained patterns, to the best of our knowledge, the problem of optimizing clean 3DGS from blurry images has not been explored yet.

In this paper, we present \ours, the first Gaussian Splatting framework designed to reconstruct 3D scenes from images blurred by camera motion.
The proposed \ours pipeline jointly optimizes the latent camera motion of blurry images and sharp 3D scenes represented by Gaussian Splatting.
Specifically, we reconstruct blurry images by accumulating 3DGS rendered images following the estimated camera motion of training views and optimize the reconstructed blurry images to converge with input blurry images.

DeblurGS improves the deblurring performance beyond existing NeRF-based deblurring approaches by leveraging the fine-detailed rendering performance of 3DGS.
Furthermore, we observe that while previous methods perform well under the ideal condition of accurately known camera poses for blurry images, they falter when blur-induced error disrupts the initial camera poses from SfM.
We analyze the behavior of 3DGS when initial camera poses are inaccurate due to blur, and propose a Gaussian Densification Annealing strategy that enables stable optimization of the sharp 3D scene.
In addition, we have found that different blurry images can be reconstructed depending on how sub-frame camera poses are sampled along the estimated camera motion trajectory. 
To achieve more accurate blur optimization, we propose the sub-frame alignment parameters, which control discrete sampling intervals on the continuous motion trajectory.

The proposed method, DeblurGS, outperforms previous methods in deblurring across all existing multi-view deblurring datasets.
In particular, DeblurGS is the only method capable of restoring high-quality sharp 3D scenes from the noisy camera poses obtained through SfM on existing datasets.
Furthermore, our experiments for real-world captured videos demonstrate the effectiveness of DeblurGS for practical applications.

In summary, our key contributions are:
\begin{compactitem}
	\item We propose DeblurGS, the first 3D Gaussian Splatting pipeline that optimizes a sharp 3D scene from motion blur images.
        \item We adopt the Gaussian Densification Annealing strategy to optimize noisy initial camera poses of input blurry images.
        \item We demonstrate the practicality of DeblurGS through experiments on field-captured, fast-moving, real-world videos.
\end{compactitem}

\section{Related Work}
\label{sec:relatedwork}

\paragraph{\textbf{Image Deblurring}}
Image deblurring is one of the fundamental tasks in the image restoration field.
The conventional deep learning approach to restore sharp images from blurry input is CNN or transformer-based supervised learning~\cite{sun2015learning, chakrabarti2016neural, wieschollek2017learning, tao2018scale, zhang2020deblurring, kupyn2019deblurgan, cho2021rethinking, chen2022simple, zamir2022restormer}.
However, these methods require a large amount of training data paired with sharp ground truth images~\cite{rim_2020_realblurdataset, rim2022realistic, nah2017deep, nah2019ntire}, and the quality of deblurring is dependent on the scale of the data.
Moreover, these methods often struggle with generalization across different conditions due to the domain gap problem, making their performance inconsistent in diverse real-world scenarios.
In contrast, our \ours framework deviates from the traditional data-driven paradigm, eliminating the need for a pre-trained network trained by large-scale datasets and inherently unaffected by a domain-gap problem.

\paragraph{\textbf{NeRF and 3D Gaussian Splatting}}

NeRF~\cite{mildenhall2020nerf} draws great attention in 3D vision fields on account of its photo-realistic view synthesis results.
The core strategy of NeRF is to optimize neural implicit representation with a differentiable volume rendering technique.
Several follow-up approaches aim to improve the rendering quality~\cite{barron2021mip,barron2022mip,jiang2022alignerf,wu2021diver,kaizhang2020nerfplusplus}, while other branches of the study are committed to mitigate the time-consuming training and rendering speed~\cite{kangle2021dsnerf,lindell2021autoint,Reiser2021ICCV,chan2022efficient,yu2021plenoctrees,wang2022r2l,chen2023mobilenerf,fridovich2022plenoxels,muller2022instant,chen2022tensorf,sun2022dvgo} of NeRF's
framework, improving the rendering speed by several orders of magnitude.
Recently, 3DGS~\cite{kerbl20233d} enhances variants of the radiance fields model and achieves both detailed reconstruction performance and real-time rendering speed.
By replacing NeRF's ray marching~\cite{maxraymarching} with rasterization that is efficient and deterministic, 3DGS achieves real-time rendering without loss of visual quality.
Our \ours is also built on a 3DGS pipeline to restore fine-grained patterns of latent sharp scenes from blurry observations and perform real-time rendering of reconstructed scenes.

\paragraph{\textbf{3D Reconstruction from Blur}}

Recently, NeRF-based deblurring approaches \cite{lee2023exblurf, wang2023badnerf, Ma_deblurnerf, Lee_2023_CVPR} attempt to reconstruct a sharp 3D scene from blurry multi-view images.
The NeRF based approaches jointly optimize the blur operation of each image with the sharp 3D scene that explains all blurry inputs.
DeblurNeRF~\cite{Ma_deblurnerf} and DP-NeRF~\cite{Lee_2023_CVPR} adopts a 2D pixel-wise blur kernel estimator, and BAD-NeRF\cite{wang2023badnerf} and ExBluRF~\cite{lee2023exblurf} directly estimate camera trajectories of each input image.
Despite impressive novel view synthesis and deblurring performance, NeRF based methods confront challenges optimizing scenes from inaccurate initial poses, which is a natural assumption considering SfM pipeline estimates erroneous camera poses if blurry views are given.
Our \ours establishes restoration of sharp 3D scenes from erroneous poses by adopting our Gaussian Densification Annealing strategy.

\section{Method}
\label{sec:method}

\begin{figure*}[t!]
    \centering
    \vspace{0.3cm}
    \includegraphics[width=0.95\textwidth]{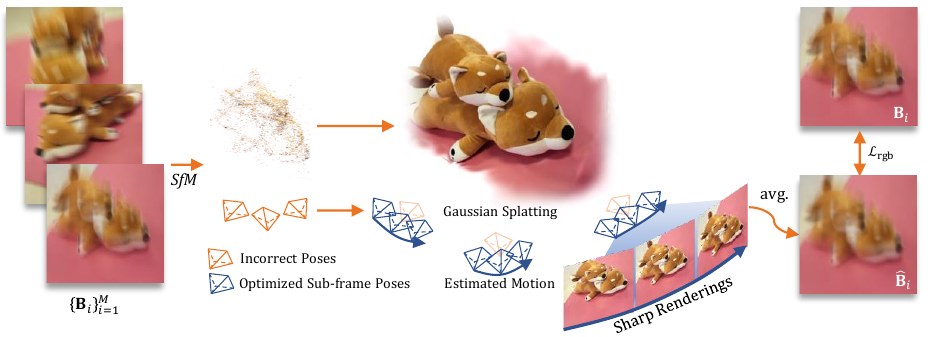}
    \caption{\textbf{Training pipeline of DeblurGS.} We simulate the physical blur operation while the camera is moving.
    In our optimization, the blurry images $\{\hat{\mathbf{B}_i}\}_{i=1}^{M}$ are reconstructed by accumulating rendered images along with the estimated camera trajectories.
    We minimize L1 loss $\mathcal{L}_\text{rgb}$ between the input blurry images $\{\mathbf{B}_i\}_{i=1}^{M}$ and reconstructed blurry images $\{\hat{\mathbf{B}_i}\}_{i=1}^{M}$ to jointly optimize the camera motion trajectories and the sharp 3D scene. }

\label{fig:method}
\end{figure*}   

We introduce \ours, a method to optimize sharp Gaussian Splatting based 3D scenes from camera motion blurred images.
Given multi-view observations which are blurry due to camera motion, our goal is to restore a sharp 3D scene.
To this end, we adopt 3DGS~\cite{kerbl20233d} as a scene representation to accomplish photo-realistic recovery of 3D scenes in fine-grained detail, and jointly optimize 3DGS with the latent camera motion.
We present a brief review of 3DGS in Sec.~\ref{sec:preliminary}.
Next, we introduce a method for blurry view synthesis by estimating camera motion and accumulating sub-frame rendering from the approximated motion in Sec.~\ref{sec:blurry_view_synthesis}.
We present the optimization process in~\cref{sec:optimization-process}, highlighting Gaussian Densification Annealing strategy for optimization from the erroneous poses.
Lastly, we derive the loss terms for the optimization in Sec.~\ref{sec:loss}.
We provide the overview of the training procedure in Fig.~\ref{fig:method}

\subsection{Preliminary: 3D Gaussian Splatting}
\label{sec:preliminary}
3D Gaussian Splatting~\cite{kerbl20233d} characterizes a 3D scene as a set of Gaussian primitives $G=\{ (\mathbf{x}_i ,\Sigma_i, c_i, \sigma_i)\}_{i=1}^{|G|}$, where $\mathbf{x}_i$, $\Sigma_i$, $c_i$, and $\sigma_i$ depict 3D coordinates, covariance matrix, RGB color and opacity of $i$-th primitive, respectively.
Omitting the intrinsic notation of the camera for simplicity, we obtain a 2D projection of 3D Gaussian represented by mean $\mathbf{x}_i$ and covariance $\Sigma_i$ given camera pose $\projmat \in SE(3)$ by:
\begin{equation}
	 \mathbf{x}_{\text{2D},i}=\projmat\mathbf{x}_i, \quad \text{and} \quad \Sigma_{\text{2D},i} = J\projmat\Sigma_i\projmat^\mathsf{T}J^\mathsf{T},
\label{eq:gaussianprojection}
\end{equation}
where $\mathbf{x}_{\text{2D},i}$ and $ \Sigma_{\text{2D},i}$ are mean and covariance of projected 2D Gaussian, repsectively, and $J$ is a Jacobian of the affine approximation of the projective transformation~\cite{surfacesplatting-zwicker}.
After projecting all Gaussians to image space, we calculate the opacity $\alpha_i=\sigma_i \exp{ \left(-\frac{1}{2} (\mathbf{x}-\mathbf{x}_{\text{2D},i})^\mathsf{T} \Sigma_{\text{2D},i}^{-1}(\mathbf{x}-\mathbf{x}_{\text{2D},i}) \right) }$ 
and obtain the pixel RGB value $\hat{c}$ at pixel $\mathbf{x}$ by $\alpha$-blending:
\begin{equation}
    \hat{c}(\mathbf{x}) = \sum_{i=1}^{|G|} \left( \prod_{j=1}^{i-1} (1-\alpha_j(\mathbf{x})) \right) \alpha_i(\mathbf{x}) c_i.
\label{eq:pixelrgb}
\end{equation}
The order of Gaussians for $\alpha$-blending is decided by sorting by distance from the camera to Gaussians.
Consequently, we represent the rendering $\rendermat_G$ of Gaussian Splatting $G$ from the pose $\projmat$ as: 
\begin{equation}
    \rendermat_G(\projmat) = \{ \hat{c}(\mathbf{x})\ \mid \mathbf{x} \in \mathcal{P} \},
\label{eq:image_render}
\end{equation}
where $\mathcal{P}$ stands for a set of all pixels on the image space.

Given multi-view images with known camera poses, the optimization of 3DGS is done by minimizing the L1 distance between the rasterized view $\rendermat_G(\projmat)$ and the ground truth image from the pose $\projmat$.
If given multi-view images are blurry caused by camera motion, na\"ively employing Gaussian Splatting training procedure results in poor 3D reconstruction as shown in \cref{fig:teaser} because minimization of the L1 norm hardly converges with blurry observation.
Our approach integrates auxiliary parameters that simulate blur operation into the original 3DGS pipeline to guide the Gaussian splattings converge to represent a sharp scene.

\begin{figure}[t!]
    \vspace{0.5cm}
    \centering
    \includegraphics[width=0.98\textwidth]{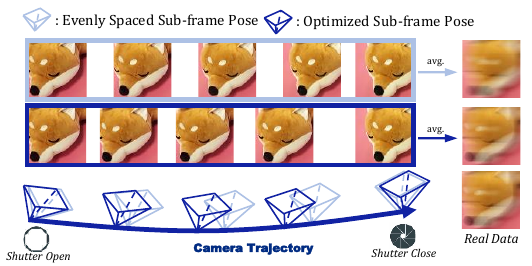}
    \caption{\textbf{Illustration of the Sub-frame Alignment Parameters.}
    With the estimated camera trajectory, the resulting blurry image changes based on the sampling intervals of sub-frame images.
    Even if the latent camera trajectory is well-optimized, the evenly sampled blurry image (\textbf{\textcolor{grayblue}{$\mathbf{1^{st}}$ row}}) does not correspond to the blurry observation of real data.
    We align the sub-frames intervals by optimizing the sub-frame alignment parameters to accumulate precise blurry image (\textcolor{navy1}{$\mathbf{2^{nd}}$ \textbf{row}}).
    }
    
    \label{fig:curve}
    
\end{figure}

\subsection{Blurry View Synthesis}
\label{sec:blurry_view_synthesis}
We aim to optimize sharp Gaussian Splatting using given motion blurred inputs.
Physically, the camera motion blur is generated from the integration of irradiance during camera movement~\cite{park2017joint} such as hand-shake or trembling.
Therefore, the acquisition of blurry image $\mathbf{B}$ is represented by an integration of irradiance for the time-varying 6-DOF camera pose $\projmat_\tau\in SE(3)$ within the exposure time $\tau\in[\tau_o,\tau_c]$:

\begin{equation}
    \mathbf{B}= \int_{\tau_o}^{\tau_c} \mathbf{I}(\projmat_\tau) d\tau \approx \frac{1}{N}\sum_{i=1}^{N}\mathbf{I}(\projmat_{\tau_i}),
    \label{eq:blur_operation_6dof}
\end{equation}
where $\mathbf{I}(\projmat_\tau)$ denotes sharp image acquired from the pose $\projmat_\tau$.
Here, we approximate the integral calculation as a finite sum of $N$ irradiance $\mathbf{I}(\projmat_{\tau_i})$, where the sub-interval timestamp $\tau_i=\tau_o +\frac{i-1}{N-1}(\tau_c-\tau_o)$.

We estimate the 6-DOF camera trajectory that explains motion blur to accurately simulate the blur operation described in~\cref{eq:blur_operation_6dof}. 
Following ExBluRF~\cite{lee2023exblurf}, we parameterize the rigid motion of the camera with the \bezier curve in the Lie algebra space $\mathfrak{se}(3)$.
However, we have found that even if we accurately estimate the camera trajectory using a \bezier curve, the blurry image is not synthesized uniquely due to differences in the sampling positions of camera poses along the trajectory, as shown in \cref{fig:curve}.
Therefore, we define the sub-frame alignment parameter $\nu= \{\nu_{i}\}_{i=1}^{N}$ 
which calibrates each camera pose $\hat{\projmat}(\nu_{i})$ on the estimated trajectory to be aligned with the latent camera poses at time $\tau_i$:
\begin{equation}
\hat{\projmat}(\nu_{i})\triangleq\projmat_{\tau_i} \quad\forall i\in \{1,2,\ldots,N\}.   
\label{eq:sub-frame-parameter}
\end{equation}
We apply the definition of the alignment parameter $\nu$ (Eq.~\ref{eq:sub-frame-parameter}) to formulate motion blur image $\mathbf{B}$ using Eq.~\ref{eq:blur_operation_6dof}:
\begin{equation}
    \mathbf{B} \approx \frac{1}{N}\sum_{i=1}^{N}\mathbf{I}(\projmat_{\tau_i})=\frac{1}{N}\sum_{i=1}^{N}\mathbf{I} (\hat{\projmat}(\nu_{i})),
\label{eq:blur_operation_camera_motion}
\end{equation}
where we abbreviate $\hat{\projmat}(\nu_{i})$ as $\hat{\projmat}_i$ for brevity. By introducing the sub-frame alignment parameter $\nu$, we overcome the challenge of non-unique blur synthesis from a single trajectory.  

We synthesize a motion blurred rendering by adopting Gaussian Splatting rasterization described in Sec. \ref{sec:preliminary} to the motion blur formulation.
Given approximate 6-DOF camera motion, we retrieve the sub-frame poses $\{\hat{\projmat}_i\}_{i=1}^{N}$ corresponding to each sub-interval $\{\tau_i\}_{i=1}^{N}$. 
We generate a blurry rendering by accumulating sharp rendering of Gaussian Splatting from the sub-frame poses.
We also apply a gamma correction function to the synthesized blurry view to correctly replicate the image acquisition process of the camera.
When a camera converts the irradiance to a digital image, a nonlinear function is applied to brighten the raw image~\cite{gonzales2008digital}.
We choose to leverage the widely used gamma correction function $\gamma(x)=x^{1/2.2}$ to the synthesized blurry view, following previous works on blur synthesis~\cite{Ma_deblurnerf,lee2023exblurf,rim_2020_realblurdataset}.
Finally, applying the rasterization process depicted in Eq.~\ref{eq:pixelrgb} to the blur operation (Eq.~\ref{eq:blur_operation_camera_motion}), we formulate the generation of blurry rendering $\hat{\mathbf{B}}$ as integration of sharp rendering $\rendermat_G(\hat{\projmat}_i)$ of the 3DGS scene $G$ from the retrieved sub-frame poses $\hat{\projmat}_i$, followed by the gamma correction function $\gamma(\cdot)$:
\begin{equation}
    \hat{\mathbf{B}}(G) = \gamma \left( \frac{1}{N}\sum_{i=1}^{N} \rendermat_G(\hat{\projmat}_i)\right).
\label{eq:blursynthesis}
\end{equation}
We minimize the distance between synthesized blurry rendering $\hat{\mathbf{B}}(G)$ and the blurry observation to optimize the latent sharp scene.

\subsection{Optimization from Inaccurate Poses}
\label{sec:optimization-process}

Given a set of $M$ blurry observations $\{\mathbf{B}_i\}_{i=1}^{M}$, our goal is to find 3DGS parameters $G$ which represent a sharp scene.
To achieve this, we define a set of camera motion parameters (Sec.~\ref{sec:blurry_view_synthesis}) to estimate camera trajectory and alignment of sub-frames.
For $i$-th camera, we assign the camera motion parameter to retrieve sub-frame poses $\{\hat{\projmat}^{(i)}_{j}\}_{j=1}^{N}$ from the estimated motion, where $\hat{\projmat}^{(i)}_{j}$ denotes the $j$-th sub-frame pose of $i$-th camera. 
We initialize the camera trajectory from the camera poses obtained by executing the SfM pipeline COLMAP~\cite{schonberger2016structure} for all blurry observations and the $N$ alignment parameter $\{\nu^{(i)}_j\}_{j=1}^{N}$ to be evenly spaced within the range of $[0,1]$.
Subsequently, the optimization process adjusts the camera trajectory and the sub-frame parameter $\nu^{(i)}$ toward latent camera motion. % as illustrated in Fig.~\ref{fig:curve}.
We accumulate sharp renderings from each sub-frame pose and reconstruct blurry rendering $\hat{\mathbf{B}}_i$ as outlined in Eq.~\ref{eq:blursynthesis}. 

\paragraph{\textbf{Gaussian Densification Annealing}}
The initial poses obtained by COLMAP are erroneous since the conventional feature matching algorithm is conducted on noisy features from blurry images.
Given the inaccurate initial poses, joint estimation of camera motion leads to the generation of the Gaussians in incorrect positions during the early stages of optimization.
Specifically, if the gradient with respect to the position of the Gaussian exceeds the densification threshold $\theta$, the Gaussian is split into two.
Consequently, the Gaussians at incorrect positions disrupt the optimization process by trying to fit the training images from the incorrect locations.
To prevent the generation of premature Gaussian splitting in an incorrect location, we employ an annealing strategy of the densification threshold $\theta$.
We gradually anneal $\theta$ down from a higher initial $\theta$, allowing for a more refined densification when the camera motion is sufficiently optimized.
The annealing strategy for $\theta$ prioritizes the camera motion's accurate optimization before the scene attempts to represent fine details through densification, thereby mitigating the disruptions caused by Gaussians at an erroneous location.
\subsection{Loss Functions}
\label{sec:loss}

\paragraph{\textbf{Reconstruction Loss}}
We employ a reconstruction loss $\mathcal{L}_{\text{rgb}}$ to converge the blurry rendering $\hat{\mathbf{B}}_i$ to the blurry observation $\mathbf{B}_i$:
\begin{equation}
    \mathcal{L}_{\text{rgb}}=\sum_{i}\norm{\mathbf{B}_i-\hat{\mathbf{B}}_i}_1.
\label{eq:rgbloss_gamma}
\end{equation}
The reconstruction loss $\mathcal{L}_\text{rgb}$ leads joint optimization of the motion trajectories, the sub-frame alignment parameters and the sharp Gaussian Splatting.

\paragraph{\textbf{Temporal Smoothness Loss}}
The reconstruction loss $\mathcal{L}_\text{rgb}$ minimizes the error between the integration of sharp renderings and the corresponding blurry observation.
However, the optimization relying solely on $\mathcal{L}_\text{rgb}$ can converge to unrealistic latent Gaussians and motion trajectories due to the ill-posedness of the deblurring problem.
To regularize the ill-posedness of the blurry view synthesis in~\cref{eq:blursynthesis}, we apply a temporal smoothness loss:

\begin{equation}
    \mathcal{L}_{\text{smooth}} = \frac{1}{N}\sum_{i,j}\norm{ \rendermat_G(\hat{\projmat}^{(i)}_{j+1}) - \rendermat_G(\hat{\projmat}^{(i)}_j)}_2,
\label{eq:smooth}
\end{equation}
where $\mathcal{L}_{\text{smooth}}$ penalizes the RGB difference of neighboring sub-frame renderings to prevent abrupt changes in the rendering of adjacent viewpoints.
We finalize our optimization objective $\mathcal{L}$ by combining $\mathcal{L}_{\text{rgb}}$ to $\mathcal{L}_{\text{smooth}}$:
\begin{equation}
    \mathcal{L} = \mathcal{L}_{\text{rgb}}+\lambda \mathcal{L}_{\text{smooth}}.
\end{equation}
We assign $\lambda=0.05$, gradually annealing down to $\lambda=0.01$ during the optimization. 
After the training procedure, we only keep sharp Gaussian Splatting $G$ and all camera motion parameters are not used in sharp view rendering at inference time.

\section{Experiment}
\label{sec:experiments}

\begin{table*}[t!]
    \vspace{0.5cm}
    \caption{Quantitative comparison of novel view synthesis.}
    \centering
    \setlength\tabcolsep{4pt} % Default value: 6pt
    \begin{tabular*}{0.9\textwidth}{@{\extracolsep{\fill}} l|ccc|ccc}
    \toprule
    & \multicolumn{3}{c}{Real Motion Blur~\cite{Ma_deblurnerf}} & \multicolumn{3}{c}{Synthetic Extreme Blur~\cite{lee2023exblurf}} \\
    Method & PSNR$\uparrow$ & SSIM$\uparrow$ & LPIPS$\downarrow$ & PSNR$\uparrow$ & SSIM$\uparrow$ & LPIPS$\downarrow$ \\
    \midrule
    3DGS~\cite{kerbl20233d} & 21.69 & 0.684 & 0.281 & 20.39 & 0.591 & 0.454 \\
    DeblurNeRF~\cite{Ma_deblurnerf} & 25.49 & 0.763 & 0.182 & 23.98 & 0.687 & 0.301 \\
    BAD-NeRF~\cite{wang2023badnerf} & 22.59 & 0.633 & 0.252 & 21.94 & 0.590 & 0.349 \\
    ExBluRF~\cite{lee2023exblurf} & 25.93 & 0.775 & 0.198 & 27.81 & 0.823 & 0.227 \\
    \textbf{Ours} & \textbf{26.28} & \textbf{0.862} & \textbf{0.086} & \textbf{30.23} & \textbf{0.890} & \textbf{0.078} \\
    \midrule
    & \multicolumn{3}{c}{ExBlur-CP\cite{lee2023exblurf}} & \multicolumn{3}{c}{ExBlur-NP\cite{lee2023exblurf}} \\
    3DGS~\cite{kerbl20233d} & 23.79 & 0.643 & 0.538 & 22.34 & 0.611 & 0.532 \\
    DeblurNeRF~\cite{Ma_deblurnerf} & 28.87 & 0.709 & 0.402 & 24.59 & 0.656 & 0.425 \\
    BAD-NeRF~\cite{wang2023badnerf} & 27.15 & 0.647 & 0.453 & 25.31 & 0.645 & 0.377 \\
    ExBluRF~\cite{lee2023exblurf} & 30.17 & 0.757 & 0.284 & 24.37 & 0.659 & 0.472 \\
    \textbf{Ours} & \textbf{31.75} & \textbf{0.806} & \textbf{0.172} & \textbf{30.11} & \textbf{0.768} & \textbf{0.192} \\
    \bottomrule
    \end{tabular*}
\label{tab:result}
\end{table*}

\subsection{Experimental Setup}

\paragraph{\textbf{Datasets}}
We perform comprehensive experiments on two real-world benchmark datasets, Real-motion-blur~\cite{Ma_deblurnerf} and ExBlur~\cite{lee2023exblurf}, and also synthetic dataset Synthetic Extreme Blur~\cite{lee2023exblurf}.
All datasets contain multiple scenes, and each scene contains 20 to 40 blurry training images with hand-shaking camera motion blur, as well as 4 to 6 sharp testing images.
For the Real-motion-blur dataset, the camera poses are obtained by COLMAP~\cite{schonberger2016structure} with a set of training and testing images.
For the ExBlur dataset, the amount of blur is more severe than the Real-motion-Blur dataset.
Each blurry image in the ExBlur dataset has its sharp pair obtained from beam-splitter~\cite{rim_2020_realblurdataset}.
The camera poses of the blurry observation are calculated from the sharp pair in the ExBlur dataset.
For the Synthetic Extreme Blur dataset also contains more severe motion blur than the Real-motion-blur dataset, and the camera pose is known by generating from Blender~\cite{blender}.
%

% Exblur-COLMAP
To better evaluate sharp 3D reconstruction capabilities in realistic scenarios, we divide the ExBlur dataset into two experimental settings based on the accuracy of initial poses and points cloud.
The first setup, referred to as ExBlur-CleanPose or \textit{ExBlur-CP}, is unchanged from the dataset's public release, which is compromised of accurate camera pose and point clouds from COLMAP with sharp image pairs.
On the other hand, ExBlur-NoisyPose or \textit{ExBlur-NP} setting exclusively utilizes only blurry images for COLMAP processing, notably excluding sharp test images as well.
To obtain the camera poses and point clouds from the inaccuracies induced by the blur, we loosen pre-defined COLMAP hyperparameters such as the re-projection error threshold and the number of inlier feature points.
We detail the modifications made for COLMAP with blurry images in the supplementary materials.
Given the significant blur that affects all images, not every image is guaranteed to be registered successfully in the process of Structure-from-Motion.
However, we proceed our experiments with a minimum of 16 images for registration, which is at least 60\% of the total images.

\def\deblurnerfresultheight{4.85cm}
\def\exblurresultheight{4.25cm}

\captionsetup[subfigure]{labelformat=empty}
\begin{figure*}[t!]
        \vspace{0.5cm}
	\centering
	\subfloat[\tiny {Blurry View}]{\includegraphics[height=\deblurnerfresultheight]{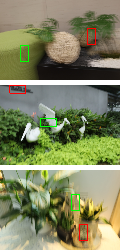}}
        \hfill
	\subfloat[\tiny Input]{\includegraphics[height=\deblurnerfresultheight]{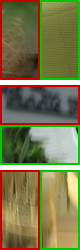}}
        \hfill
        \subfloat[\tiny 3DGS]{\includegraphics[height=\deblurnerfresultheight]{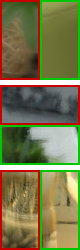}}
        \hfill
        \subfloat[\tiny DeblurNeRF]{\includegraphics[height=\deblurnerfresultheight]{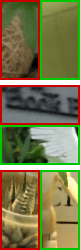}}
        \hfill
        \subfloat[\tiny BAD-NeRF]
        {\includegraphics[height=\deblurnerfresultheight]{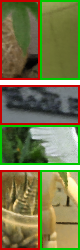}}
        \hfill
        \subfloat[\tiny ExBluRF]{\includegraphics[height=\deblurnerfresultheight]{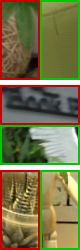}}
        \hfill
        \subfloat[\tiny \textbf{\ours}]{\includegraphics[height=\deblurnerfresultheight]{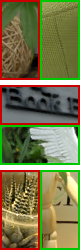}}

\caption{\textbf{Qualitative Comparison of Deblurring on Real Motion Blur}~\cite{Ma_deblurnerf}. Note that blurry views in Real Motion Blur do not have their ground-truth pairs.}
\label{fig:result_deblurnerf}
\end{figure*}

\captionsetup[subfigure]{labelformat=empty}
\begin{figure*}[t!]
    
        \centering
	\subfloat[\tiny{Blurry View}]{\includegraphics[height=\exblurresultheight]{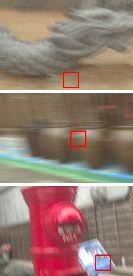}}
        \hfill
	\subfloat[\tiny Input]{\includegraphics[height=\exblurresultheight]{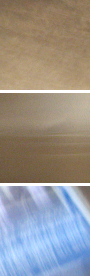}}
        \hfill
        \subfloat[\tiny 3DGS]{\includegraphics[height=\exblurresultheight]{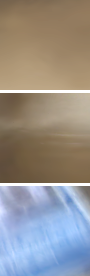}}
        \hfill
        \subfloat[\tiny DeblurNeRF]{\includegraphics[height=\exblurresultheight]{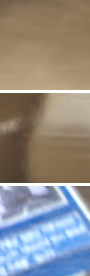}}
        \hfill
        \subfloat[\tiny BAD-NeRF]{\includegraphics[height=\exblurresultheight]{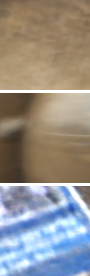}}
        \hfill
        \subfloat[\tiny ExBluRF]
        {\includegraphics[height=\exblurresultheight]{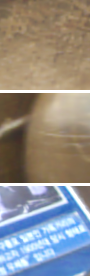}}
        \hfill
        \subfloat[\textbf{\tiny \ours}]{\includegraphics[height=\exblurresultheight]{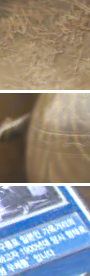}}
        \hfill
        \centering
        \subfloat[\tiny{GroundTruth}]{\includegraphics[height=\exblurresultheight]{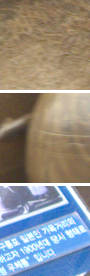}}
        
\caption{\textbf{Qualitative Comparison of Deblurring on ExBlur-CP}~\cite{lee2023exblurf}. The camera poses are initialized from COLMAP~\cite{schonberger2016structure} with sharp pairs corresponding to each blurry observation. }
\label{fig:result_exblur}
\end{figure*}

\paragraph{\textbf{Implementation Details}}
Our \ours pipeline is built upon the official implementation of 3DGS~\cite{kerbl20233d}.
We set the number of sub-frames for blurry rendering synthesis to $N=21$, the same as ExBluRF~\cite{lee2023exblurf} for a fair comparison. 
Additionally, we use the $9$-th order \bezier curve in our camera trajectory estimation.
We use Adam optimizer~\cite{2015-kingma} with default configuration, and all the learning rate schema regarding the 3DGS model is unchanged from its built-in setting.
For the camera motion parameters, we schedule the learning rate of $1\times10^{-2}$ for the translation part of the camera trajectory, $1\times10^{-3}$ for the rotation, and $3\times10^{-3}$ for the alignment parameter $\nu$, all of which decay exponentially to be halved for every 15k iterations.
Each scene is optimized for 150k iterations using a single NVIDIA GeForce RTX 4090 GPU. 

\vspace{0.5cm}

\def\exblurcolmapwidth{0.95\textwidth}
\begin{figure}[t!]
    \vspace{0.7cm}
    \centering % Center the entire figure within the text

    \newcommand{\verticalcaption}[1]{%
        \begin{minipage}{0.2cm} % Adjust the width of the minipage for the rotated caption
        \centering
        \rotatebox[origin=c]{90}{\scriptsize{#1}} % Rotate subcaption
        \end{minipage}%
    }

    % Subfigure 1
    \verticalcaption{Blurry View}
    \begin{minipage}[c]{\exblurcolmapwidth}
        \includegraphics[width=\linewidth]{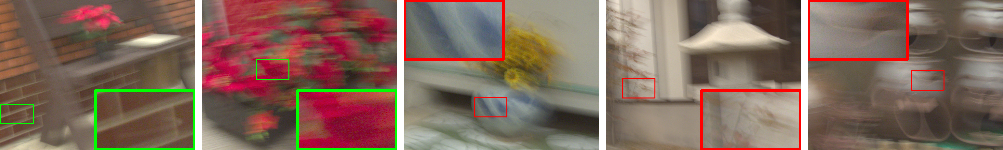} % Adjust width as needed
    \end{minipage}\hfill % Adjust spacing between the subfigures as needed

    % Subfigure 2
    \verticalcaption{3DGS}
    \begin{minipage}[c]{\exblurcolmapwidth}
        \includegraphics[width=\linewidth]{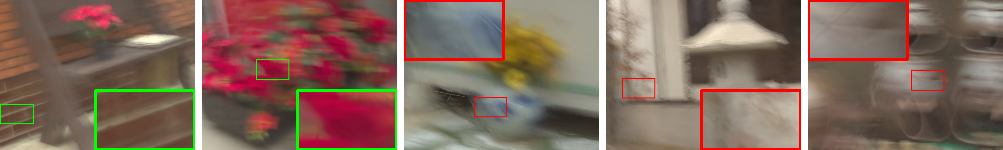}
    \end{minipage}\hfill

    % Subfigure 3
    \verticalcaption{DeblurNeRF}
    \begin{minipage}[c]{\exblurcolmapwidth}
        \includegraphics[width=\linewidth]{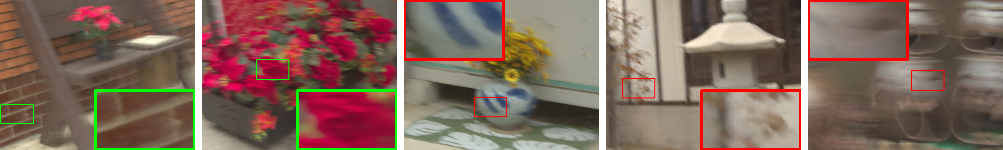}
    \end{minipage}\hfill

    % Subfigure 4
    \verticalcaption{BAD-NeRF}
    \begin{minipage}[c]{\exblurcolmapwidth}
        \includegraphics[width=\linewidth]{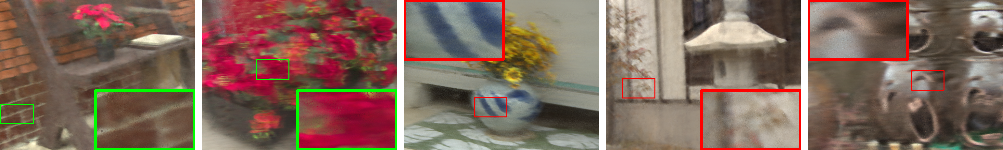}
    \end{minipage}\hfill

    % Subfigure 5
    \verticalcaption{ExBluRF}
    \begin{minipage}[c]{\exblurcolmapwidth}
        \includegraphics[width=\linewidth]{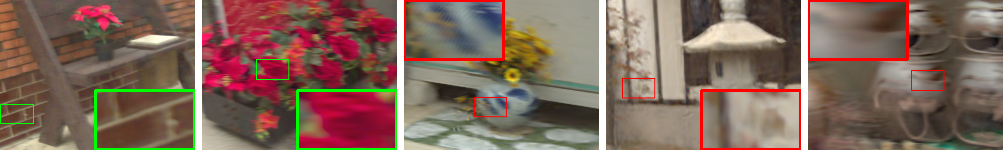}
    \end{minipage}\hfill

    % Subfigure 6
    \verticalcaption{\textbf{DeblurGS}}
    \begin{minipage}[c]{\exblurcolmapwidth}
        \includegraphics[width=\linewidth]{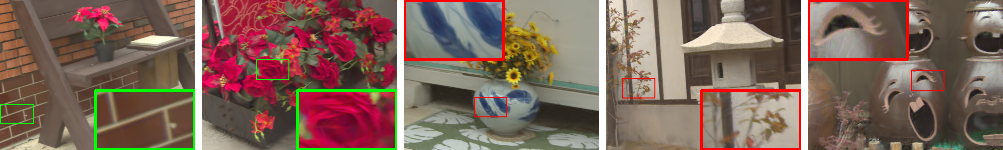}
    \end{minipage}\hfill

    % Subfigure 7
    \verticalcaption{Ground Truth}
    \begin{minipage}[c]{\exblurcolmapwidth}
        \includegraphics[width=\linewidth]{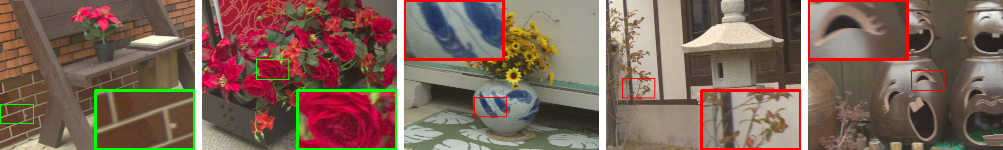}
    \end{minipage}

    \caption{\textbf{Qualitative comparison of sharp 3D rendering on ExBlur-NP}~\cite{lee2023exblurf}. The camera poses are initialized from COLMAP~\cite{schonberger2016structure} with blurry observations only.}
    \label{fig:result_difficult}
\end{figure}

\subsection{Results}
We compare the novel view synthesis performance to vanilla 3D Gaussian Splatting~\cite{kerbl20233d}, DeblurNeRF~\cite{Ma_deblurnerf}, BAD-NeRF~\cite{wang2023badnerf}, and ExBluRF~\cite{lee2023exblurf}.
DeblurNeRF approximates the blur kernel on the 2D plane and optimizes with Neural Radiance Fields~\cite{mildenhall2020nerf}.
BAD-NeRF and ExBluRF jointly estimate the 6-DOF camera trajectory as a form of a linear~\cite{wang2023badnerf} and B\`ezier curve~\cite{lee2023exblurf} in the Lie algebra ($\mathfrak{se}(3)$) space, respectively.

\paragraph{\textbf{Evaluation Setup}}

The performance of novel view synthesis is evaluated by measuring PSNR, SSIM~\cite{hore2010image}, and LPIPS~\cite{zhang2018perceptual} between rendering and sharp test images.
For all settings, since test poses are misaligned due to the joint optimization of 6-DOF camera motion, we perform the post-optimization process to align the test image poses, following previous 3D reconstruction works on pose-free settings~\cite{yen2020inerf,SCNeRF2021,bian2022nopenerf,lin2021barf,fu2023colmapfree} and blurry-image settings~\cite{wang2023badnerf,lee2023exblurf}.
In the ExBlur-NP setup, we fix the training camera poses and generate sharp renderings from these viewpoints to conduct COLMAP's image registration and acquire poses for test images. 
After the test image registration, we align the test image poses by the post-optimization process similar to the evaluation process for the other setups.
Note that the scene parameters such as NeRF, voxels and 3DGS remain frozen during post-optimization, and the purpose of the test-time optimization is strictly limited to factoring out the pose misalignment of test view images.

\newcommand{\cmark}{\textcolor{green}{\ding{51}}} % checkmark
\newcommand{\xmark}{\textcolor{red}{\ding{55}}} % X mark

\begin{table}[t]
    \vspace{0.7cm}
   \caption{\textbf{Ablation study.} We describe the ablation studies on each element of proposed method on "Camellia" and "Stone Lantern" scenes in ExBlur dataset with noisy camera pose setup. We ablate the effectiveness of the temporal smoothness loss $\mathcal{L}_{\text{smooth}}$, Gaussian densification annealing strategy, sub-frame alignment parameters $\nu$, and the gamma correction $\gamma(\cdot)$.}
    
    \centering
    \scalebox{0.90}{
    \begin{tabular}{cccc|ccc|ccc}
        \toprule[1.0pt]
         % \multicolumn{5}{c|}{} & \multicolumn{6}{c}{Scene} \\
         % \midrule
         \multicolumn{4}{c|}{Method} & \multicolumn{3}{c|}{Camellia} & \multicolumn{3}{c}{Stone Lantern} \\
         {$\mathcal{L}_{\text{smooth}}$} & Anneal $\theta$ & learn $\nu$ & $\gamma(\cdot)$ & PSNR$\uparrow$ & SSIM$\uparrow$ & LPIPS$\downarrow$ & PSNR$\uparrow$ & SSIM$\uparrow$ & LPIPS$\downarrow$  \\
        \midrule
        \xmark & \xmark & \xmark & \xmark & 26.78 & 0.686 & 0.215 & 24.92 & 0.688 & 0.277 \\
        \xmark & \cmark & \cmark & \cmark & 27.22 & 0.653 & 0.252 & 26.20 & 0.716 & 0.275 \\
        \cmark & \xmark & \cmark & \cmark & 28.12 & 0.680 & 0.243 & 24.35 & 0.699 & 0.269 \\
        \cmark & \cmark & \xmark & \cmark & 28.12 & 0.675 & 0.253 & 28.20 & 0.772 & 0.234 \\
        \cmark & \cmark & \cmark & \xmark & 27.35 & 0.699 & 0.225 & 25.48 & 0.729 & 0.256 \\
        \midrule
        \cmark & \cmark & \cmark & \cmark & \textbf{28.97} & \textbf{0.711} & \textbf{0.189} & \textbf{29.07} & \textbf{0.779} & \textbf{0.194} \\
        % \textbf{\ours (Ours)} & \textbf{26.16} & \textbf{0.871} & \textbf{0.072} & \textbf{31.41} & \textbf{0.799} & 0.170 \\
        \bottomrule[1.0pt]
    \end{tabular}
    }
    
    \label{tab:ablation}
\end{table}

\paragraph{\textbf{Evaluation on Deblurring}}
As shown in \cref{tab:result}, quantitative results show that our \ours outperforms previous methods across all metrics.
Comparing with the results from vanilla 3DGS, we can evident that previous NeRF-based deblurring methods are capable of restoring sharp images under conditions of moderate blur or when camera poses computed from sharp images.
While deblurring performance differences among methods on the Real-motion-blur datasets are minimal, this gap significantly increases with the synthetic and ExBlur-CP datasets, which feature more severe motion blur. 
Notably, the proposed \ours outperforms ExBluRF by a large margin, even though ExBluRF is explicitly designed to handle extreme motion blur.
The qualitative results in \cref{fig:result_deblurnerf} and \cref{fig:result_exblur} visually support the performance gain presented in \cref{tab:result}.
Interestingly, we have found that \ours can render images even sharper than their ground-truth images in certain scenes of ExBlur dataset.

However, the results from accurate initial poses represent an ideal scenario, because the accurate camera poses cannot be obtained by SfM from blurry images.
We demonstrate that the proposed \ours is the only deblurring method capable of being applied in real-world scenarios, as evidenced by the difference in deblurring performance between ExBlur-NP and ExBlur-CP setups. 
BAD-NeRF incorporates a bundle-adjustment mechanism in their optimization, so the performance drops are relatively smaller compared to other NeRF-based methods.
However, \cref{fig:result_difficult} shows that only our method can restore scenes that are close to the ground truth sharp images.
\ours maintains its plausible deblurring performance even with the inaccurate pose initialization and performs almost equally to ExBluRF with the clean pose initialization in terms of PSNR, while exhibiting better performance in SSIM and LPIPS.

\subsection{Ablation}
We present an ablation study on two scenes, Camellia and Stone Lantern, in the Exblur-NP dataset to verify and analyze the effectiveness of each component of the proposed approach. 
\cref{tab:ablation} indicates that each component plays a crucial role in sharp scene reconstruction and the highest performance is achieved when all elements are combined.

\paragraph{\textbf{Temporal Smoothness Loss}}
We find that ignoring the smoothness between adjacent sub-frame renderings leads to rendering with jittering artifacts, and degrades the metric as shown in \cref{tab:ablation}.
Without the smooth transitions between neighboring sub-frame renderings, we observe that the accumulation of unrealistic renderings with heavy artifacts reconstructs the blurry observation perfectly, due to the ill-posedness of the deblurring problem.

\paragraph{\textbf{Gaussian Densification Annealing}}
We observe that maintaining a constant threshold for Gaussian densification during the observation process generates Gaussians at inaccurate locations before camera motion converges.
Subsequently, the Gaussians in erroneous positions attempt to fit the observation and cause floating artifacts, which deteriorate the quality of reconstruction as detailed in \cref{tab:ablation}.

\paragraph{\textbf{Alignment Parameter}}
We detect a slight drop in effectiveness as recorded in \cref{tab:ablation} when employing uniformly sampled poses from an estimated trajectory to synthesize blur instead of sampling sub-frame poses from learnable alignment parameters.
Due to the lack of a complete formulation of physical motion, convergence to latent camera motion is challenging.

\paragraph{\textbf{Gamma Correction}}
Omitting a gamma correction function on synthesized blur overlooks a process occurring in real cameras, resulting in areas naturally dark appearing unexpectedly brighter.
Consequently, as illustrated in~\cref{tab:ablation}, elevating the scene's brightness leads to a substantial decline in performance.

\captionsetup[subfigure]{labelformat=empty}
\def\customheight{5.8cm}
\def\arrowgap{-0.3cm}
\vspace{0.7cm}
\begin{figure*}[t!]
    \centering
    \subfloat[]{\includegraphics[height=\customheight]{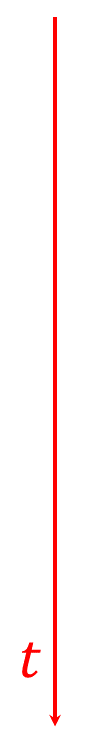}}
    \hfill
    \hspace{\arrowgap}
    \subfloat[Blurry Frames]{\includegraphics[height=\customheight]{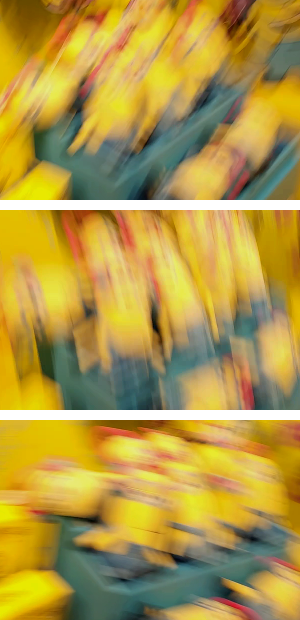}}
    \hfill
    \subfloat[Sharp Rendering]{\includegraphics[height=\customheight]{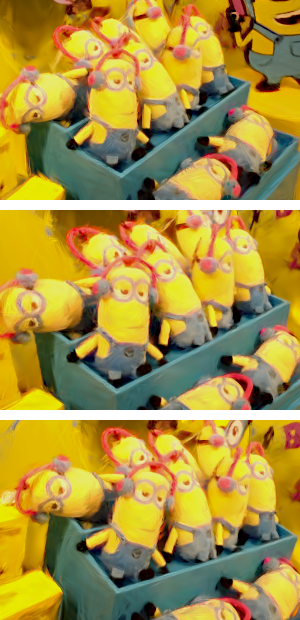}}
    \hfill
    \hspace{-0.4cm}
    \subfloat[]{\includegraphics[height=\customheight]{fig/result_custom/t.png}}
    \hfill
    \hspace{\arrowgap}
    \subfloat[Blurry Frames]{\includegraphics[height=\customheight]{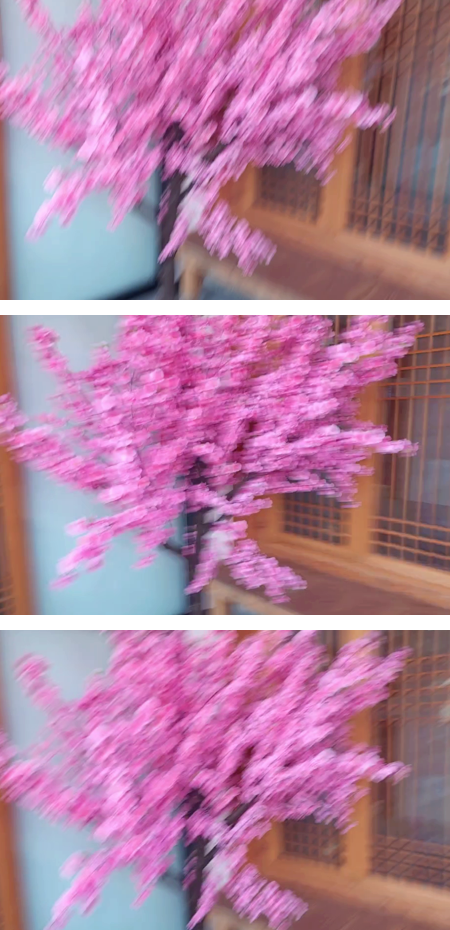}}
    \hfill
    \subfloat[Sharp Rendering]{\includegraphics[height=\customheight]{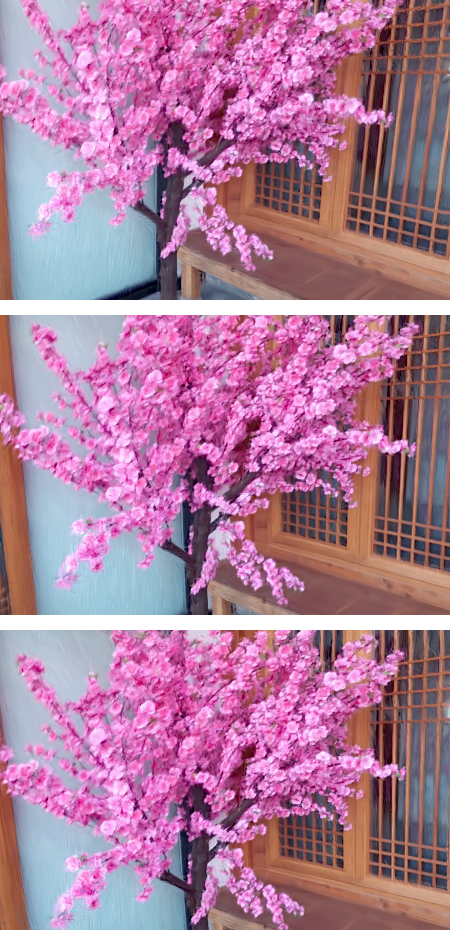}}
    \caption{\textbf{Sharp scene reconstruction from field-captured video.} We record the real-world scene as a video with smartphone, swiftly to capture the whole environment within 6 seconds. Our method successfully reconstructs sharp scene with a real-world blurry video.}
\label{fig:custom}
\end{figure*}

\subsection{Real-World Video Analysis}
To evaluate the practical applicability of our \ours method, we conduct experiments using videos captured with a fast-moving smartphone.
Using a Galaxy Ultra 22 device, we capture the surrounding environment at 30 fps for 4 to 6 seconds.
We deploy our framework by taking each frame of the video as input and rendering a sharp view after the optimization process.
As exhibited in \cref{fig:custom}, our method reconstructs sharp scenes even starting with field-captured video with blurry frames.
\ours effectively reconstructs sharp details in real-world environments since our method robustly operates on erroneous COLMAP~\cite{schonberger2016structure} initialization, showcasing the effectiveness for practical applications.

\section{Conclusion}
\label{sec:conclusion}

In this paper, we present \ours, a method for reconstructing a sharp 3D scene from a collection of motion blurred images.
We simulate the camera motion to synthesize a blurry view, and optimize 3D Gaussian Splatting by minimizing distance between given blurry observation and generated blur.
With our Gaussian Densification Annealing strategy, the camera motion converges to the latent camera movement, even initialized from noisy camera poses, the natural outcome of SfM from blurry observations.
The capability to optimize from imprecise poses underscores the practicality of our framework, showcasing the successful deblurring of videos captured with smartphones.
\ours outperforms all existing methods in the sharp 3D scene reconstruction task, achieving state-of-the-art in both experimental and practical settings.

\section{Acknowledgement}
This work was supported in part by the SNU-NAVER Hyperscale AI Center, IITP grants [No. 2021-0-01343, Artificial Intelligence Graduate School Program (Seoul National University), No.2021-0-02068, and No.2023-0-00156], the NRF grant [No.2021M3A9E4080782] funded by the Korean government (MSIT).

\bibliographystyle{splncs04}
\bibliography{main}
\end{document}